\documentclass[conference]{IEEEtran}
\IEEEoverridecommandlockouts
\usepackage{subfigure} 
\usepackage{cite}
\usepackage{balance}
\usepackage{amsmath,amssymb,amsfonts}
\usepackage{algorithmic}
\usepackage{graphicx}
\usepackage{textcomp}
\usepackage{xcolor}
\usepackage[utf8x]{inputenc}
\usepackage{mathrsfs}
\usepackage{tabularx}  
\usepackage{bm}
\usepackage{color}

\def\BibTeX{{\rm B\kern-.05em{\sc i\kern-.025em b}\kern-.08em
    T\kern-.1667em\lower.7ex\hbox{E}\kern-.125emX}}

\begin{document}

\title{PointLocalization: Real-Time, Environmentally-Robust\\ 3D LiDAR
Localization}
\author{Yilong Zhu$^{}$, Bohuan Xue$^{}$, Linwei Zheng$^{}$, Huaiyang Huang, Ming Liu$^{}$, Rui Fan$^{}$\\
$^{}$Robotics and Multi-Perception Laborotary, Robotics Institute,\\  Hong Kong University of Science and Technology, Clear Water Bay, Hong Kong SAR, China.\\
Email: \{yzhubr, bxueaa, lzhengad, hhuangat, eelium, eeruifan\}@ust.hk
\vspace{-2.0em}
}

\maketitle

\begin{abstract}

Localization, or position fixing, is an important problem in robotics research. In this paper, we propose a novel approach for long-term localization in a changing environment using 3D LiDAR. We first create the map of a real environment using GPS and LiDAR. Then, we divide the map into several small parts as the targets for cloud registration, which can not only improve the robustness but also reduce the registration time. PointLocalization allows us to fuse different kinds of odometers, which can optimize the accuracy and frequency of localization results. We evaluate our algorithm on an unmanned ground vehicle (UGV) using LiDAR and a wheel encoder, and obtain the localization results at more than 20 Hz after fusion. The algorithm can also localize the UGV in a 180-degree field of view (FOV). Using an outdated map captured six months ago, this algorithm shows great robustness, and the test results show that it can achieve an accuracy of 10 cm. PointLocalization has been tested for a period of more than six months in a crowded factory and has operated successfully over a distance of more than 2000 km.  
\end{abstract}

\section{Introduction}
\label{sec.introduction}

The deployment of autonomous vehicles has been increasing rapidly since Google first launched their self-driving car project in 2009 \cite{fan2017real, fan2018real}. One of the key aspects in robotics research is localization, particularly in the application of autonomous driving and unmanned ground vehicles (UGVs) \cite{liu2018mobile}, where vehicles operate in special areas. Localization of the vehicle is important, so that the path planning module can send it  navigation commands. Localization is the basis of navigation. 

According to Paul \cite{groves2013principles}, a navigation module contains two parts: position fixing, which involves comparing features at the current location, such as landmarks, way-points, and maps; and dead reckoning, which means measuring the change in  either the position or the velocity, such as visual odometry and wheel odometry. 

 UGVs travel along a certain route, such as in baggage delivery, and navigate in factories and mining sites. In these situations, we can use a map to guide the vehicle’s travel along the route. Many different types of sensors can be used for localization \cite{fan2019key}. When position fixing is used for navigation, a global positioning system (GPS) is one of the most common sensors, which can locate nearly any position in the world. However, it highly relies on satellites, which means it does not work well in the city where signals may be blocked. To overcome the shortfalls, GPS can be coupled with an inertial navigation system (INS) by using a Kalman filter. This increases both the accuracy and robustness. However, the cost of this system is high and it can only operate for about one minute when the GPS signal has been lost. 
 
Localization or position fixing by LiDAR or camera requires a fixed map or landmarks. Simultaneous localization and mapping (SLAM) can be used to obtain a fixed map, and it has wide usage in many scenarios, such as in sweeping robots. Oriented FAST and rotated BRIEFSLAM (ORB-SLAM) \cite{mur2015orb} is a SLAM framework based on a camera, which shows great robustness in pose estimation. It also provides a localization function. However, the ORB features change during light variances, and this change may cause failure, for example, mapping takes place in the morning and location occurs in the evening. It is difficult to replace ORB-SLAM with other position fixing systems, like GPS, when we navigate a fixed map using a camera. This method shows great accuracy in visual odometry, and the localization function also shows great potential for use in industrial areas. 
 
 LiDAR has recently become a fundamental sensor for autonomous driving because it measures the distance to a target by illuminating the targets. It does not depend on the light, unlike a camera. Zhang \cite{zhang2014loam} proposed a method called LiDAR odometry and mapping (LOAM), which uses LiDAR to get the point cloud map, and uses the distance of corner and planar points to optimize the target of the Levenberg–Marquardt method. LOAM shows real-time mapping and state-of-art accuracy. However, this method cannot eliminate the cumulative error, and we cannot locate using the map. Thrun \cite{levinson2010robust} uses the LiDAR reflectivity to build a map, and this method shows an accuracy of around 10 cm. However, this method only works when there are a sufficient number of landmarks on the ground, and it is affected by snow and rain. 
 
To solve the challenges mentioned above, we build a point cloud map to eliminate the illumination changes and we use GPS to constrain our map. This step reduces the drift over a long distance. In order to make the localization more robust, we use other kinds of odometry to predict our correspondence point on the map, and the odometry also achieves a faster updating speed than LiDAR. 

The remainder of this paper is structured as follows. Section II illustrates the related work, and connects other method with our approach. Section III presents the sensor and platform we use, as well as a mapping method based on graph optimization. Section IV presents the PointLocalization method.  The results on the KITTI dataset and the real-world environment are discussed in Section V. Section VI illustrates the experimental results and outlines future works.

\section{Related Work}
In this section, we provide an overview of the existing SLAM methods and compare PointLocalization with other localization methods. 

As previously mentioned, ORB-SLAM \cite{mur2015orb} provides a localization method that uses ORB features \cite{rublee2011orb} to calculate visual odometry, but it is sensitive to the change of brightness and features in the environment. To solve this problem, \cite{ding2018laser} first uses LiDAR to build a previous map based on the LiDAR map geometry information, and proposes a hybrid bundle adjustment framework to correct the current pose. Wolcott and Eustice \cite{wolcott2014visual} also use a LiDAR map as prior knowledge of the environment; however, it uses the ground map generated by LiDAR to match the camera results.

Another method to locate using a camera is proposed in \cite{burki2016appearance} \cite{burki2018map}, in which presented an online landmark selection method using an unsupervised algorithm makes a selection according to how likely these landmarks are, for distributed long-term visual localization systems. They also proposed a map management system for long-term lifelong localization. For autonomous driving, localization is still a challenging problem for industrial usage, because it highly relies on the features of the camera, such as dynamic range and photosensitivity, which can affect the robustness of visual localization in different conditions. 

Koide et al. \cite{koide2018portable} uses graph SLAM to fusion the LiDAR and GPS messages. They use a normal distributions transform (NDT) \cite{biber2003normal} algorithm to scan matching, and a GPS as the constraint of the pose graph. They also combine the NDT registration algorithm with angular velocity data provided by the LiDAR using an unscented Kalman filter (UKF) to localize. Pfrunder and Egger \cite{pfrunder2017real}, \cite{egger2018posemap} proposed a new SLAM method called CSIRO SLAM (C-SLAM), which uses IMU and LiDAR. It calculates surfels by their spatial relation from the LiDAR, and minimizes the error of the transform from surfels matching and the data of acceleration and rotation from the IMU to get current pose. They also offer a localization method, called CSIRO localization (C-LOC), which also uses surfel matching and fuses the data from the IMU to get current localization results. 

One problem for localization is how to get a pre-built map with high-quality. Many researchers use the loop closure method to reduce the drift during map generation. LEGO-LOAM \cite{shan2018lego} uses the trajectory to estimate the loop closure, then it calculates the root mean square error between the current LiDAR frame and the key frame, that was acquired in the past as the loop closure condition. ORB-SLAM also uses loop closures to reduce the accumulation error. It detects the bag of words (BOW) \cite{mur2014fast}, which describes  the kind of object in a certain frame as the feature vector, and if the trajectory of the camera comes to the same place, will show the same vector value. Another problem that occurs when using LiDAR to locate is the real-time performance. For example, a 64-beam LiDAR will generate 2.2 million points per second, makes it hard to calculate the result in real time.

In this paper, we present a method based on GPS to build a map without drift. In the mapping part, we use LOAM as the estimation of the sensor trajectory, and we set up a posed map to reduce the accumulation errors. In the posed map section, we add a GPS constraint and loop closure constraint to get a map without drift, and this GPS-based mapping method algorithm can achieve real-time performance.

\section{Experimental Set-Up and G-LOAM}
\subsection{Sensor Settings}
\subsubsection{Industrial Personal Computer (IPC)}
Our IPC is equipped with an Intel Core i7 7700 CPU (3.6 GHz) and a 16 GB  of RAM. The IPC works on Ubuntu 16.04 with a Robot operating system (ROS) Kinetic version. 

\subsubsection{3D Lidar}
RS-LiDAR-16 Lidar in our experiments is used. RS-LiDAR-16 is a mechanical LiDAR, which has a 30{\textdegree} vertical field of angle and 360{\textdegree} horizontal field of angle. RS-LiDAR-16 has 16 different laser channels, and each channel works at a 10 Hz rotation rate. 

  \subsubsection{Encoder}
 On our test vehicle, we use a Hall effect sensor as our wheel speed sensor, which is used to measure the magnitude of a magnetic field. We build a 36T gear that can rotate the wheels. When the vehicle moves forward, the Hall effect sensor will generate an electric signal, according to the magnetic field changes generated by the rotating gear. We calculate the travel distance by counting the amount of pulse from the Hall effect sensor. We also obtain the yaw angle according to the different electric signals generated by the left and right wheels. 
 
In this paper, we define the distance between the left and right wheels as $L$. $d$ is the traveling distance of the wheel. We can obtain the traveling distance of the wheel center using $d_\text{center} = \frac{d_\text{left}+d_\text{right}}{2}$. The turnning angle is $\phi = \frac{d_\text{left}+d_\text{right}}{2}$. From the former position 
$
\begin{bmatrix} X_i, & Y_i, & \theta_i \end{bmatrix}^\top
$, we can calculate the UGV position using:
\begin{equation}
T_{i+1}=\begin{bmatrix}
X_i\\Y_i\\ \theta_i
\end{bmatrix}
+
\begin{bmatrix}
d_\text{center}\cos\theta_i\\
d_\text{center}\sin\theta_i\\\phi
\end{bmatrix}.
\label{a}
\end{equation}

\subsection{G-LOAM}

In this section, we introduce our graph-based SLAM method with GPS constraint, which is called G-LOAM. Firstly, when using LiDAR to map, we must use a different point cloud registration method as the front end. With this step, we can get LiDAR odometry. Researchers usually use iterative closest points (ICP) \cite{holz2015registration} or NDT as the front end registration method.   NDT is widely used because it does not need an accurate initial position and it has stable time consumption. We propose a mapping method for which the registration part is based on LOAM. LOAM is more accurate and less time-consuming than other registration algorithms, because it only matches the features instead of whole LiDAR scan. Furthermore, a loop closure method is usually used to optimize the map, LEGO-LOAM uses ICP as the loop closure constraint. We add a GPS constraint to the loop closure constraint, in order to eliminate the accumulation error of LOAM. The general framework for graph optimization (G2o) \cite{kummerle2011g} is the basic frame to solve our problems.

 A graph-based method \cite{grisetti2010tutorial} is one of the most successful approaches to solve SLAM problems. In the graph, we use the edge to represent different constraints, such as the GPS measurement to the pose of LiDAR. The goal of the graph is to minimize the error of all measurements. 

Let $\bm{p_{k}}$ be the LiDAR node, where $\bm{Z_{k}}$ $\bm{H_{k}}$ respectively represent the mean of the constraint of $\bm{p_{k}}$ and the information matrix of the current constraint of $\bm{p_{k}}$. We can also define the error function $\bm{e_k}(\bm{p_{k},\bm{Z_k})}$ between $\bm{p_{k}}$ and the observation by using the edge to represent it. We can use a Eq. 2 to describe this optimization problem: 
\begin{equation}
F(p)=\sum_{k=0}^N\bm{e_k}(\bm{p_{k}},\bm{Z_k})^\top\bm{H_{k}}\bm{e_k}(\bm{p_{k}},\bm{Z_k}).
\label{b}
\end{equation}  
Following Eq. 2, we select the Levenberg-Marquardt (LM) algorithm to solve this problem. The LM algorithm is more effective than Gauss-Newton algorithm, because the LM algorithm sets up a trusting region where the non-linear approximation is valid. 
\begin{figure}[t!]
	\centering
	\includegraphics[width=0.40\textwidth]{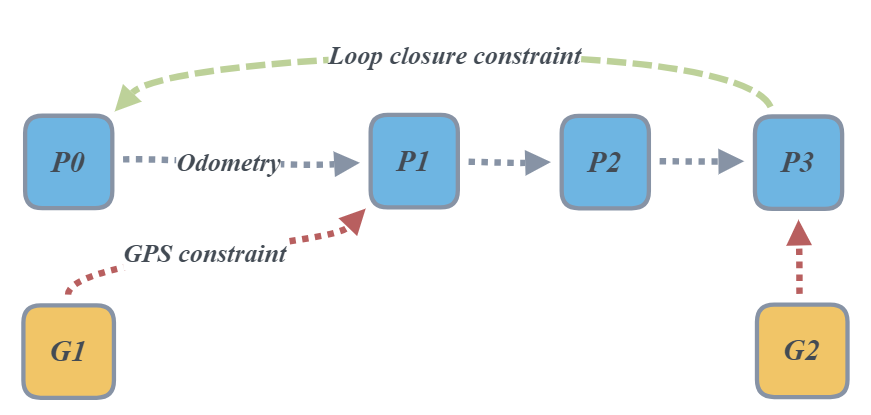}
	\caption{The proposed pose graph structure}
	\vspace{-1.5em}
\end{figure}

Fig.~\ref{a} shows the structure of our graph, we set the $G_k$ as a fixed node generated by GPS data and $P_k$ represents the LiDAR pose in SE(3). This graph has 3 constraints: the odometry constraint generated by LiDAR odometry, the GPS constraint given by the static transformation between the LiDAR and GPS antenna, and the loop closure constraint given by the ICP algorithm. Fig.~\ref{b} illustrates a map got by using this algorithm.


\section{PointLocalzation}
\subsection{PointLocalization Overview}

\begin{figure}[t!]
	\centering
		{
	\includegraphics[width = 0.32\textwidth]{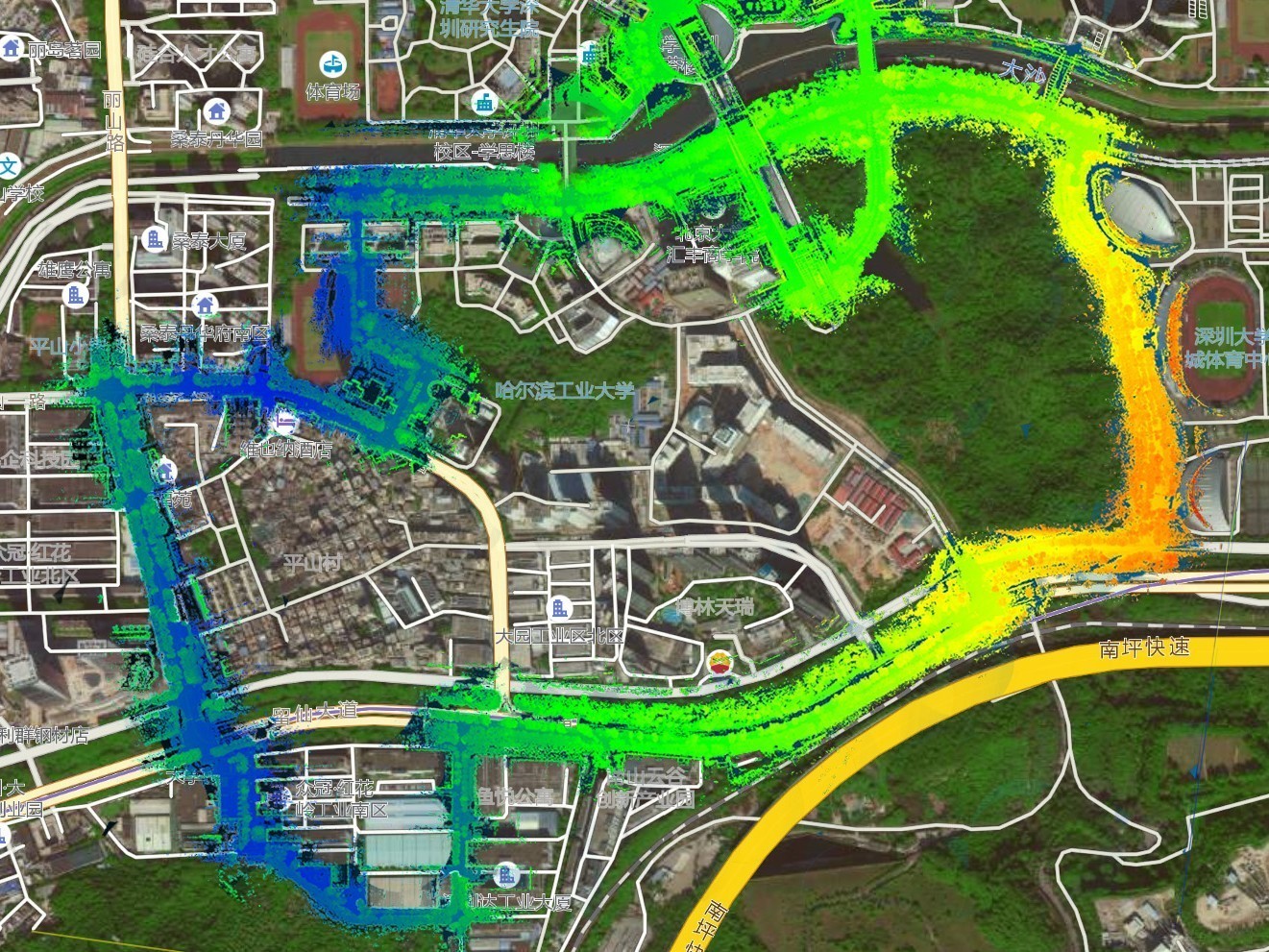}
	}
	\caption{The result of mapping on campus.}
	\label{map}
	\vspace{-1.5em}
\end{figure}

The point localization algorithm requires two sensor inputs. The first is LiDAR, which can give us the current surrounding environment. With this information, we can reduce the accumulation error. The other sensor can provide any kind of odometry. We use Hall effect sensor to provide odometry; however, it can use any other odometry that has an updated rate above 10 Hz. The sensor should be stable and not be affected by the environment change. 
\subsection{PointLocalization Algorithm}
Before introducing PointLocalization, we should finish sparsification of our point cloud generated by G-LOAM. We use a voxel grid filter to reduce the size and the overlapped area of our map. After the sparsification step, we can get a map with faster access speed and smaller storage size; the storage size is only 8 MB for a 0.5 km$\times$0.5 km map. 
\begin{figure*}[t!]
	\centering
	\subfigure[]
	{
	\includegraphics[width=0.23\textwidth]{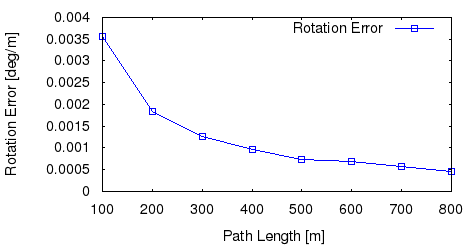}
	}
	\subfigure[]
	{
	\includegraphics[width=0.23\textwidth]{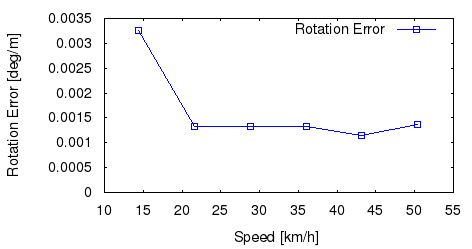}
	}
	\subfigure[]
	{
	\includegraphics[width=0.23\textwidth]{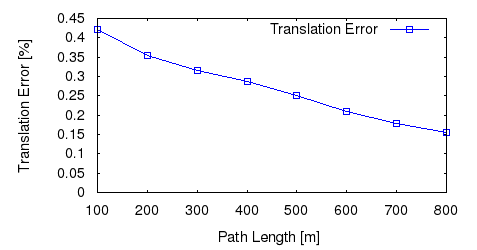}
	}
	\subfigure[]
	{
	\includegraphics[width=0.23\textwidth]{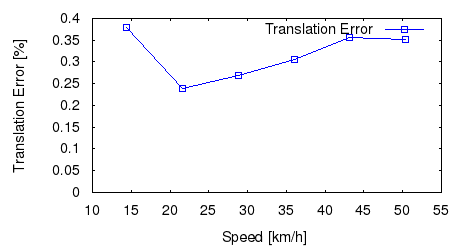}
	}

	\caption{The test result of KITTI 00. (a) Result of the trajectory of PointLocalization and ground truth. (b) The calculation time distribution. (c), (d), (e) and (f) The rotation and translation error in different distance and speed.}
	\label{c}
	\vspace{-1.9em}
\end{figure*}
After we receive the message from LiDAR and the odometry, we calculate the transformation between the last LiDAR localization result $T_{k-1}^L$ and the current received odometry $T_{k}^O$. We name it $T_{k}^I$. If we receive a LiDAR input, $T_{k}^I$ will be one prediction for our current pose $T_{k}^L$. We also use another way to predict our pose, which regard as our LiDAR momentum $(T_{k-1}^L)^T \times T_{k-2}^L$.
In this step we can get a rough prediction of $T_{k}^L$, which is $T_{k}^I$ and $(T_{k-1}^L)^T \times T_{k-2}^L$. Then we use the rough predict of $(T_{k}^L)$ to get the region of interest (ROI) on the map we obtained previously. We set the minimum ROI on the map with a radius of one meter, this step can help us increase the speed for matching the LiDAR scan points with the map.

In the last step, we get the target point cloud (ROI map) and source point cloud (LiDAR scan). We can use ICP to calculate the pose of LiDAR and formulate the problem as follows:
\begin{equation}
\bm{e}_i=
\begin{cases}
\bm{p}_i - (\bm{R}\bm{\acute{p}}_i + \bm{t})& \left\|(\bm{p}_i-\bm{\acute{p}}_i)\right\|_2<\delta\\
0& \left\|(\bm{p}_i-\bm{\acute{p}}_i)\right\|_2\geq\delta
\end{cases},
\end{equation}
and
\begin{equation}
\mathop{\min}_{\bm{R,t}}J=\frac{1}{2}\sum_{i=1}^{n}{\|(\bm{p}_i- (\bm{R}\bm{\acute{p}}_i+\bm{t}))\|_2^2}.
\label{d}
\end{equation}

We define the error of Eq. 3; $\bm{e}_i$  is the error between the LiDAR points ($\bm{p}_i$) and the ROI points in the map ($\bm{\acute{p}}_i$), $\bm{R}$ is the current rotation, $\bm{t}$ is the current transformation. In this step, we set a threshold $\delta$ to prevent the influence, caused by dynamic objects, such as a truck or pedestrians. 
Eq. 4 is the target function in ICP algorithm to get the current pose of LiDAR, where $\bm{R}$ and $\bm{t}$ should be adjusted to make sure the current estimation of the pose has the minimal $J$.


The last step is to fuse the odometry result and the LiDAR result to get a high frequency of localization result. Usually, the EKF approach can only take measurements from the current state into account, which is a problem with delayed measurements. In this part, we use the steady-state approximation of EKF (SSKF) \cite{valls2018design}. The covariance is assumed to be constant. In this method, we can get the trade-off for the delayed measurements between the accuracy of the EKF and the runtime of the SSKF; this method will solve the problem of the long calculation time, caused by the ICP algorithm.
\section{Experimental Results}

\begin{figure}[t!]
	\centering
		{
	\includegraphics[width = 0.38\textwidth]{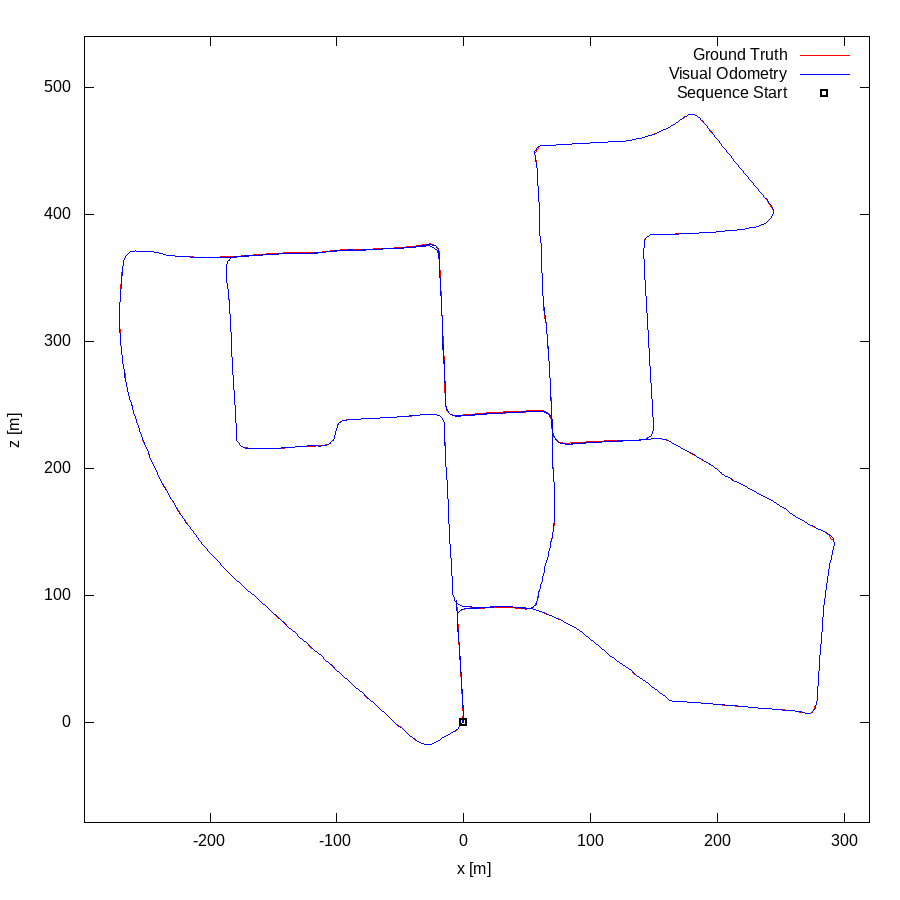}
	}
	\caption{The localization result of KITTI 00.}
	\label{e}
	\vspace{-1em}
\end{figure}

We test the G-LOAM and PointLocalization on the KITTI dataset. This experiment is performed on the IPC mentioned before. We choose the first dataset to test our algorithm. 

In the mapping part, we transform the ground truth as the GPS constraint for mapping, and we get a map with less drift. When running the PointLocalization algorithm, we use 16 beams of a 64-beam LiDAR to achieve the same standard as our testing vehicle. We also use LiDAR odometry to replace the Hall effect sensor, because KITTI does not have wheel odometry for tests.

\begin{table}  

\caption{Translation Error Result from KITTI 00}  

\begin{tabular}{ p{4cm}<{\centering} | p{4cm}<{\centering} } 
\hline  
\textbf{Error measure (m)} & \textbf{Value}   \\ 
\hline  
Mean & 0.227894   \\  
Median  & 0.141765  \\  
RMSE  & 0.324986  \\  
Std  & 0.231690  \\ 
\hline  
\end{tabular}  
\vspace{-0.2cm}  
\end{table}

Fig. 3 and Fig. 4 show the testing result; Fig. 4 is the result after comparing the localization result with the KITTI ground truth. Fig. 3(a) and Fig. 3(b) are the rotation errors in the test. From these two figures, we can get the conclusion that this algorithm works well in a 20-50 km/h speed range. Fig. 3(c) and Fig. 3(d) show the translation errors with path length and speed. We can find that this localization method does not have an accumulated error caused by odometry, and at the speed of about 20km/h, it shows excellent behavior. This is fit for a vehicle working in an urban area. Table 1 shows the total error result on KITTI 00. This shows that in the urban environment, PointLocalization can reach decimeter localization accuracy.

\subsection{Test Result on Real Data}

\begin{figure*}[t!]
	\centering
	\subfigure[]
	{
		\includegraphics[width=0.28\textwidth]{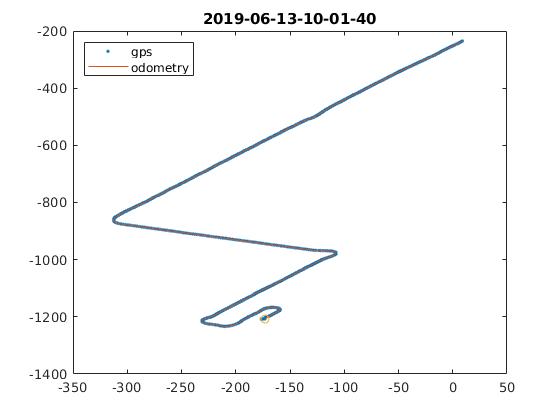}
	}
	\subfigure[]
	{
		\includegraphics[width=0.28\textwidth]{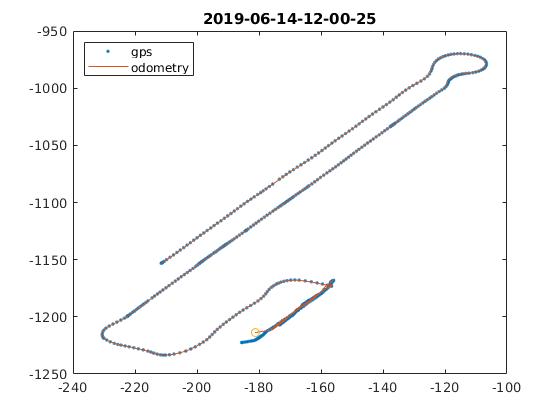}
	}
	\subfigure[]
	{
	\includegraphics[width=0.28\textwidth]{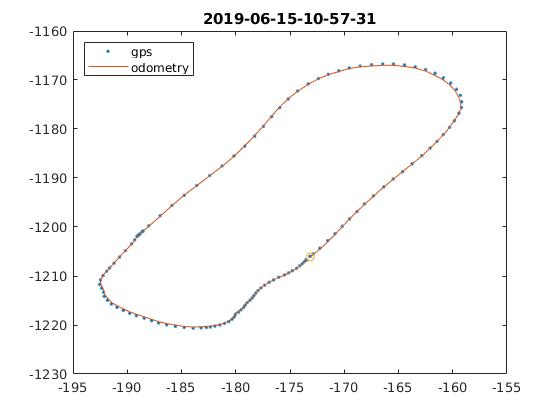}
	}
	\subfigure[]
	{
	\includegraphics[width=0.28\textwidth]{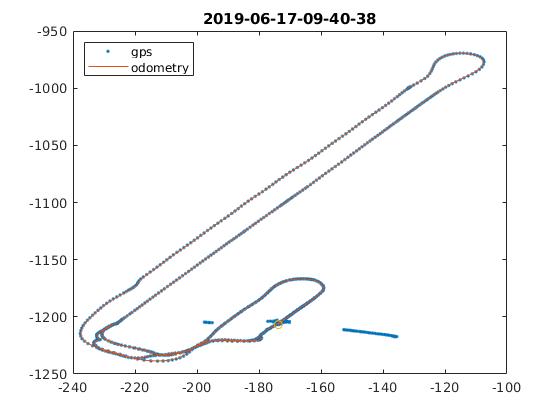}
	}
	\subfigure[]
	{
		\includegraphics[width=0.28\textwidth]{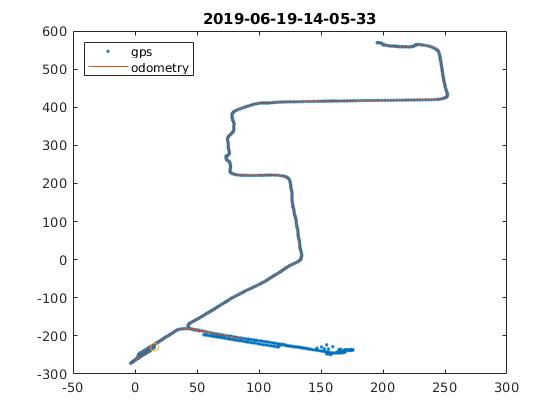}
	}
	\subfigure[]
	{
		\includegraphics[width=0.28\textwidth]{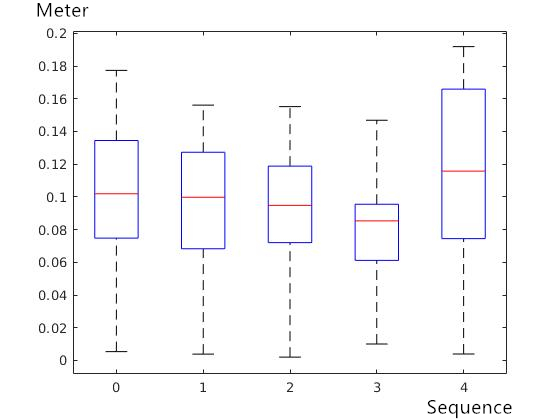}
	}
	\caption{Test result between RTK GPS and PointLocalization result. (a), (b), (c), (d) and (e) The testing result in the city at different times. (f) The accuracy of data from the results in (a0, (b), (c). (d) and (e).}
	\label{f}
	\vspace{-1.9em}
\end{figure*}

\begin{figure}[t!]
\centering
	{
	\includegraphics[width=0.35\textwidth]{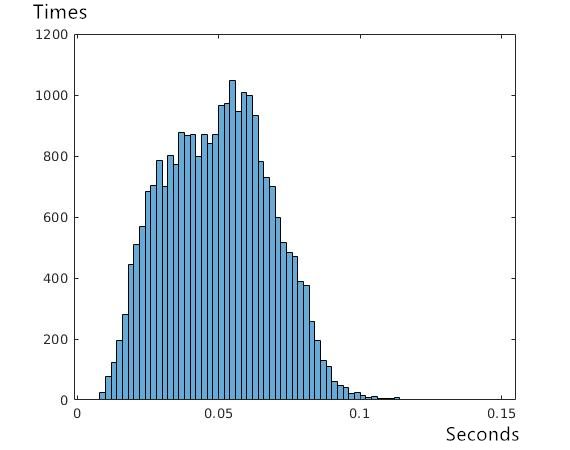}
	}
		\caption{The calculation time of our test.}
	\label{g}
	\vspace{-1.1em}
\end{figure}

We test this algorithm in our operating environment using an electric UGV for which the designed speed is 25 km/h. The trial is carried out at an industrial estate that has many trucks and people. 

 In this industrial estate, in order to verify the environmental robustness, we use a map that was captured six months ago, in January 2019. To test as many situations as possible, we record the data over a week. The results are shown in Fig. 5. Then, we specially choose the data record in the morning and afternoon, because they are the busiest periods of the estate. Fig. 5 (a)-(e) represent the RTK-GPS route and PointLocalization route. We can see from Fig. 5 (b), (d) and (e) that there are some failure situations of RTK-GPS, which are represented by the blue dot around the route. The reasons for these failure in the tests are
 \begin{itemize}
 \item{\textbf{Communications.}}
  We choose to use a continuously operating reference stations (CORS) system, which uses a fixed station to calculate the error message. Since we may lose the error message caused by unstable communications, the position we obtain is also unstable.
 \item{\textbf{Signal obstructions and Multipath.}} 
 Our vehicle works in the environment with lots of high buildings and trucks. The building will cause signal obstructions, and metallic objects located near the antenna can cause signal reflection. These elements may cause the GPS to fail during our test.
 \end{itemize}

In these experiments, we also use the RTK-GPS as the ground truth to verify the accuracy of our algorithm. (f) is the error we compare with RTK-GPS in status 4 and the RMS (66.7\%) is less than 2 cm. The error is calculated in the form: $e(t) = \sqrt{e_x(t)^2+e_y(t)^2}$.  $e_x(t)$ and $e_y(t)$ is the error between the RTK-GPS and PointLocalization in the $x$ and $y$ axis, respectively. 

PointLocalization can reach the decimeter-level localization and this result shows in the estate environment. From the test result in (f), we get the minimum error of 0.0136 m, the average error is 0.0973 m. It is better than C-LOC \cite{pfrunder2017real} whose minimum Euclidean norm is 0.096 m, the mean is 0.995 m.

The real-time demand is very important for localization. Fig. 6 represents the calculation time distribution for our algorithm. The LiDAR works at the speed of 10 Hz and we find that most of the calculation finish in 0.1 seconds. This algorithm is much faster than C-LOC \cite{egger2018posemap}, where they use a 16 beams LiDAR, and reach a 4 Hz update rate.
\begin{figure}[t!]
	\centering
		{
	\includegraphics[width = 0.32\textwidth]{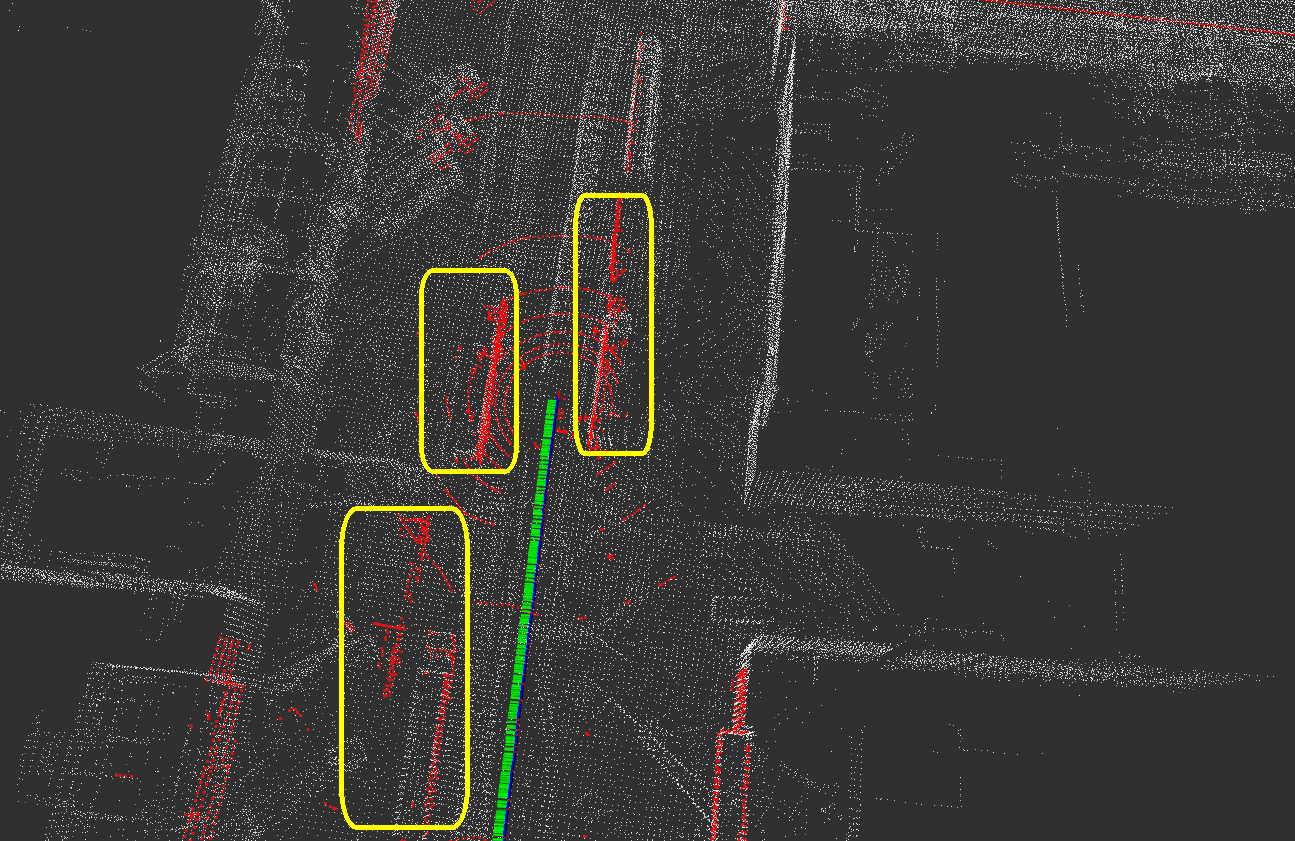}
	}
	\caption{Yellow boxes indicate trucks driving pass our vehicle.}
	\label{h}
	\vspace{-1em}
\end{figure}
Robustness is another aspect to evaluate the localization algorithm. We test it in an environment crowded with trucks. As shown in Fig. 7, the current LiDAR scan is blocked by three trucks. If we use a local map to match the scanned results, it will cause failure. With the prediction and ROI map, PointLocalization does not have a significant influence on the localization result.

In the practical usage, it is not always possible to use 360{\textdegree} LiDAR, two LiDARs, which are installed in front and back of vehicle to reduce cost and avoid obstacles. One defect of this system is that it only has a 180{\textdegree} horizontal field of angle because it is blocked by the UGV body, and the other defect is that this kind of LiDAR is usually installed at a low position. Under these circumstances, we also test this algorithm on the front LiDAR.

Fig. 8 is the localization test on an open square with a front LiDAR, where we obtain localization in an open area with only half a constraint on the map.

\begin{figure}[t!]
	\centering
		{
	\includegraphics[width = 0.32\textwidth]{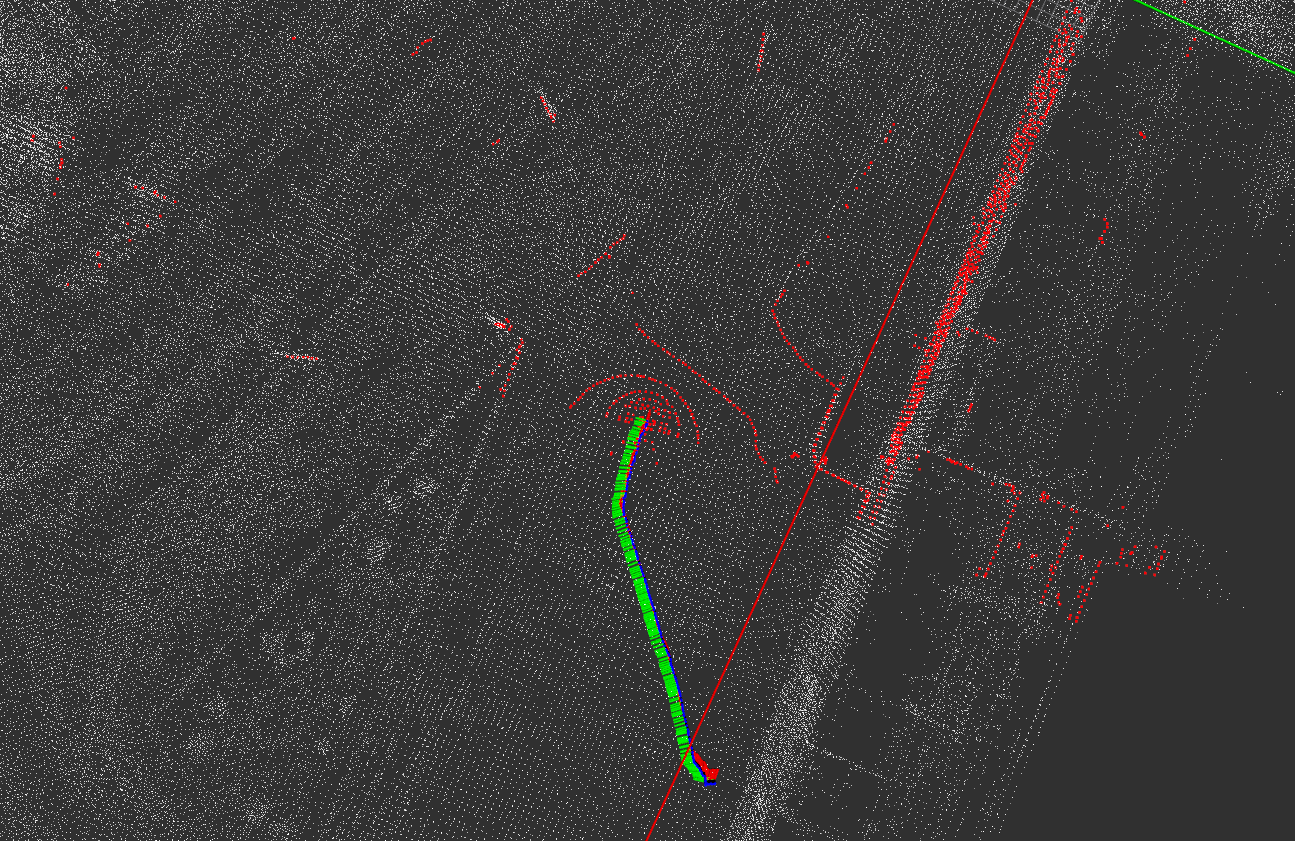}
	}
	\caption{Localization in wide environment with a 180 {\textdegree} horizontal FOV.}
	\label{i}
	\vspace{-1.9em}
\end{figure}

\label{sec.conclusion_future_work}

\section{Conclusion and Future Work}

This paper presented a method of localization based on LiDAR and any kind of odometry. The PointLocalization method reached localization to a decimeter-level in real time, and shows environmental robustness, which enables the vehicle to locate in an urban environment without installing GPS. It can also work well in crowded places like industrial estate. This method only needs one LiDAR and it will not be disturbed by light variance and surrounding buildings. These features show greater advantages than GPS and visual-based methods. 

PointLocalization requires a good prior map generated by G-LOAM, which uses GPS constraint to eliminate accumulation errors. To achieve real-time and environmentally robust localization, we use odometry and the former pose of LiDAR to estimate when the next scan should appear, then we generate the ROI map based on the current scan. In the point cloud registration part, to reduce the influence of dynamic objects, we drop the points which have large errors. 

In the experiment part, we test our algorithm on two environments: the KITTI urban environment, and a crowed industrial estate. From the KITTI test  results, we find that our algorithm can work at a high speed with high accuracy. In the  estate test with a distance of 2000 km, we find this algorithm shows great advantages: higher accuracy and faster speed than the other methods. 

In the future, we plan to improve the accuracy of the proposed method. LiDAR only has a measurement error of about 2 cm and RTK-GPS also has the same accuracy under the best conditions, so it is possible for PointLocalization to reach centimeter-level accuracy. In the future work, we can also use PointLocalization to update the map in a dynamic area, such as a port or parking lot.

\section*{Acknowledgment}
This work was supported by the National Natural Science Foundation of China, under grant No. U1713211, the Research Grant Council of Hong Kong SAR Government, China, under Project No. 11210017, No. 21202816, and the Shenzhen Science, Technology and Innovation Commission (SZSTI) under grant JCYJ20160428154842603, awarded to Prof. Ming Liu.

\bibliographystyle{IEEEtran}

\begin{thebibliography}{10}
\providecommand{\url}[1]{#1}
\csname url@samestyle\endcsname
\providecommand{\newblock}{\relax}
\providecommand{\bibinfo}[2]{#2}
\providecommand{\BIBentrySTDinterwordspacing}{\spaceskip=0pt\relax}
\providecommand{\BIBentryALTinterwordstretchfactor}{4}
\providecommand{\BIBentryALTinterwordspacing}{\spaceskip=\fontdimen2\font plus
\BIBentryALTinterwordstretchfactor\fontdimen3\font minus
  \fontdimen4\font\relax}
\providecommand{\BIBforeignlanguage}[2]{{%
\expandafter\ifx\csname l@#1\endcsname\relax
\typeout{** WARNING: IEEEtran.bst: No hyphenation pattern has been}%
\typeout{** loaded for the language `#1'. Using the pattern for}%
\typeout{** the default language instead.}%
\else
\language=\csname l@#1\endcsname
\fi
#2}}
\providecommand{\BIBdecl}{\relax}
\BIBdecl

\bibitem{fan2017real}
R.~Fan and N.~Dahnoun, ``Real-time implementation of stereo vision based on
  optimised normalised cross-correlation and propagated search range on a
  gpu,'' in \emph{2017 IEEE International Conference on Imaging Systems and
  Techniques (IST)}.\hskip 1em plus 0.5em minus 0.4em\relax IEEE, 2017, pp.
  1--6.

\bibitem{fan2018real}
\BIBentryALTinterwordspacing
R.~Fan, ``Real-time computer stereo vision for automotive applications,'' Ph.D.
  dissertation, University of Bristol, 2018. [Online]. Available:
  \url{https://ethos.bl.uk/OrderDetails.do?uin=uk.bl.ethos.761099}
\BIBentrySTDinterwordspacing

\bibitem{liu2018mobile}
Y.~Liu, R.~Fan, B.~Yu, M.~J. Bocus, M.~Liu, H.~Ni, J.~Fan, and S.~Mao, ``Mobile
  robot localisation and navigation using lego nxt and ultrasonic sensor,'' in
  \emph{2018 IEEE International Conference on Robotics and Biomimetics
  (ROBIO)}.\hskip 1em plus 0.5em minus 0.4em\relax IEEE, 2018, pp. 1088--1093.

\bibitem{groves2013principles}
P.~D. Groves, \emph{Principles of GNSS, inertial, and multisensor integrated
  navigation systems}.\hskip 1em plus 0.5em minus 0.4em\relax Artech house,
  2013.

\bibitem{fan2019key}
\BIBentryALTinterwordspacing
R.~Fan, J.~Jiao, H.~Ye, Y.~Yu, I.~Pitas, and M.~Liu, ``Key ingredients of
  self-driving cars,'' \emph{arXiv:1906.02939}. [Online]. Available:
  \url{http://eusipco2019.org/program/satellite-workshops/}
\BIBentrySTDinterwordspacing

\bibitem{mur2015orb}
R.~Mur-Artal, J.~M.~M. Montiel, and J.~D. Tardos, ``Orb-slam: a versatile and
  accurate monocular slam system,'' \emph{IEEE transactions on robotics},
  vol.~31, no.~5, pp. 1147--1163, 2015.

\bibitem{zhang2014loam}
J.~Zhang and S.~Singh, ``Loam: Lidar odometry and mapping in real-time.'' in
  \emph{Robotics: Science and Systems}, vol.~2, 2014, p.~9.

\bibitem{levinson2010robust}
J.~Levinson and S.~Thrun, ``Robust vehicle localization in urban environments
  using probabilistic maps,'' in \emph{2010 IEEE International Conference on
  Robotics and Automation}.\hskip 1em plus 0.5em minus 0.4em\relax IEEE, 2010,
  pp. 4372--4378.

\bibitem{rublee2011orb}
E.~Rublee, V.~Rabaud, K.~Konolige, and G.~R. Bradski, ``Orb: An efficient
  alternative to sift or surf.'' in \emph{ICCV}, vol.~11, no.~1.\hskip 1em plus
  0.5em minus 0.4em\relax Citeseer, 2011, p.~2.

\bibitem{ding2018laser}
X.~Ding, Y.~Wang, D.~Li, L.~Tang, H.~Yin, and R.~Xiong, ``Laser map aided
  visual inertial localization in changing environment,'' in \emph{2018
  IEEE/RSJ International Conference on Intelligent Robots and Systems
  (IROS)}.\hskip 1em plus 0.5em minus 0.4em\relax IEEE, 2018, pp. 4794--4801.

\bibitem{wolcott2014visual}
R.~W. Wolcott and R.~M. Eustice, ``Visual localization within lidar maps for
  automated urban driving,'' in \emph{2014 IEEE/RSJ International Conference on
  Intelligent Robots and Systems}.\hskip 1em plus 0.5em minus 0.4em\relax IEEE,
  2014, pp. 176--183.

\bibitem{burki2016appearance}
M.~B{\"u}rki, I.~Gilitschenski, E.~Stumm, R.~Siegwart, and J.~Nieto,
  ``Appearance-based landmark selection for efficient long-term visual
  localization,'' in \emph{2016 IEEE/RSJ International Conference on
  Intelligent Robots and Systems (IROS)}.\hskip 1em plus 0.5em minus
  0.4em\relax IEEE, 2016, pp. 4137--4143.

\bibitem{burki2018map}
M.~B{\"u}rki, M.~Dymczyk, I.~Gilitschenski, C.~Cadena, R.~Siegwart, and
  J.~Nieto, ``Map management for efficient long-term visual localization in
  outdoor environments,'' in \emph{2018 IEEE Intelligent Vehicles Symposium
  (IV)}.\hskip 1em plus 0.5em minus 0.4em\relax IEEE, 2018, pp. 682--688.

\bibitem{koide2018portable}
K.~Koide, J.~Miura, and E.~Menegatti, ``A portable 3d lidar-based system for
  long-term and wide-area people behavior measurement,'' 2018.

\bibitem{biber2003normal}
P.~Biber and W.~Stra{\ss}er, ``The normal distributions transform: A new
  approach to laser scan matching,'' in \emph{Proceedings 2003 IEEE/RSJ
  International Conference on Intelligent Robots and Systems (IROS 2003)(Cat.
  No. 03CH37453)}, vol.~3.\hskip 1em plus 0.5em minus 0.4em\relax IEEE, 2003,
  pp. 2743--2748.

\bibitem{pfrunder2017real}
A.~Pfrunder, P.~V. Borges, A.~R. Romero, G.~Catt, and A.~Elfes, ``Real-time
  autonomous ground vehicle navigation in heterogeneous environments using a 3d
  lidar,'' in \emph{2017 IEEE/RSJ International Conference on Intelligent
  Robots and Systems (IROS)}.\hskip 1em plus 0.5em minus 0.4em\relax IEEE,
  2017, pp. 2601--2608.

\bibitem{egger2018posemap}
P.~Egger, P.~V. Borges, G.~Catt, A.~Pfrunder, R.~Siegwart, and R.~Dub{\'e},
  ``Posemap: Lifelong, multi-environment 3d lidar localization,'' in \emph{2018
  IEEE/RSJ International Conference on Intelligent Robots and Systems
  (IROS)}.\hskip 1em plus 0.5em minus 0.4em\relax IEEE, 2018, pp. 3430--3437.

\bibitem{shan2018lego}
T.~Shan and B.~Englot, ``Lego-loam: Lightweight and ground-optimized lidar
  odometry and mapping on variable terrain,'' in \emph{2018 IEEE/RSJ
  International Conference on Intelligent Robots and Systems (IROS)}.\hskip 1em
  plus 0.5em minus 0.4em\relax IEEE, 2018, pp. 4758--4765.

\bibitem{mur2014fast}
R.~Mur-Artal and J.~D. Tard{\'o}s, ``Fast relocalisation and loop closing in
  keyframe-based slam,'' in \emph{2014 IEEE International Conference on
  Robotics and Automation (ICRA)}.\hskip 1em plus 0.5em minus 0.4em\relax IEEE,
  2014, pp. 846--853.

\bibitem{holz2015registration}
D.~Holz, A.~E. Ichim, F.~Tombari, R.~B. Rusu, and S.~Behnke, ``Registration
  with the point cloud library: A modular framework for aligning in 3-d,''
  \emph{IEEE Robotics \& Automation Magazine}, vol.~22, no.~4, pp. 110--124,
  2015.

\bibitem{kummerle2011g}
R.~K{\"u}mmerle, G.~Grisetti, H.~Strasdat, K.~Konolige, and W.~Burgard, ``g 2
  o: A general framework for graph optimization,'' in \emph{2011 IEEE
  International Conference on Robotics and Automation}.\hskip 1em plus 0.5em
  minus 0.4em\relax IEEE, 2011, pp. 3607--3613.

\bibitem{grisetti2010tutorial}
G.~Grisetti, R.~Kummerle, C.~Stachniss, and W.~Burgard, ``A tutorial on
  graph-based slam,'' \emph{IEEE Intelligent Transportation Systems Magazine},
  vol.~2, no.~4, pp. 31--43, 2010.

\bibitem{valls2018design}
M.~I. Valls, H.~F. Hendrikx, V.~J. Reijgwart, F.~V. Meier, I.~Sa, R.~Dub{\'e},
  A.~Gawel, M.~B{\"u}rki, and R.~Siegwart, ``Design of an autonomous racecar:
  Perception, state estimation and system integration,'' in \emph{2018 IEEE
  International Conference on Robotics and Automation (ICRA)}.\hskip 1em plus
  0.5em minus 0.4em\relax IEEE, 2018, pp. 2048--2055.

\end{thebibliography}
\balance

\end{document}